# SVM-based Multiview Face Recognition by Generalization of Discriminant Analysis

Dakshina Ranjan Kisku, Hunny Mehrotra, Jamuna Kanta Sing, and Phalguni Gupta

*Abstract*—Identity verification of authentic persons by their multiview faces is a real valued problem in machine vision. Multiview faces are having difficulties due to non-linear representation in the feature space. This paper illustrates the usability of the generalization of LDA in the form of canonical covariate for face recognition to multiview faces. In the proposed work, the Gabor filter bank is used to extract facial features that characterized by spatial frequency, spatial locality and orientation. Gabor face representation captures substantial amount of variations of the face instances that often occurs due to illumination, pose and facial expression changes. Convolution of Gabor filter bank to face images of rotated profile views produce Gabor faces with high dimensional features vectors. Canonical covariate is then used to Gabor faces to reduce the high dimensional feature spaces into low dimensional sub-spaces. Finally, support vector machines are trained with canonical sub-spaces that contain reduced set of features and perform recognition task. The proposed system is evaluated with UMIST face database. The experiment results demonstrate the efficiency and robustness of the proposed system with high recognition rates.

*Keywords*—Biometrics, Multiview face Recognition, Gabor wavelets, LDA, SVM.

## I. INTRODUCTION

BIOMETRIC authentication systems primarily use in security scenarios such as, sensitive public/private area surveillance and access control. On the use of authentication systems largely in public and private places for access control and security, face recognition/verification attracted the attention of vision researchers for several years after the first time when face detection and recognition boomed up in biometric. Many significant approaches have proposed for face recognition based on 2D and 3D images. Authentication of users with frontal view faces extensively studied [1], [2], [3], [8], [9], [17]. However, face recognition with multiview faces is still be a challenged for invariant and robust recognition. In multiview face recognition system, difficulties occurred due to non-linear representation in feature spaces. To minimize the demerits that often occur in multiview faces, a global representation to non-linear feature spaces is necessary. In addition, variations in facial expressions, lighting conditions, occlusions, un-suitable environments, and unwanted noises, affine distortions, clutter, etc, may also give some bad impact on the overall performance of face recognition accuracy. To ensure robust recognition of multiview faces with high recognition rate, some remarkable strategies proposed in [1-3], [9-10], [18]. However, multiview face recognition still is a versatile and most challenging authentication problem in terms of different viable constraints.

In face recognition algorithms, appearance-based approach uses holistic texture features and make a reduced set feature vector that can be applied to either on the whole-face or on the divided block in a face image. Sub-space based many face recognition techniques discussed so far includes PCA [4 -7], LDA [7], ICA [9], Kernel PCA [10], etc. Principal component analysis (PCA) uses to reduce the high dimensionality feature space to the smaller intrinsic dimensionality of feature space.

The main idea of using PCA [6] for face recognition is to express the large 1-D vector of pixels constructed from 2-D facial image into the compact principal components of the feature space. This can be called eigenspace projection. Eigenspace is calculated by identifying the eigenvectors of the covariance matrix derived from a set of facial images (vectors). PCA based approaches typically include two phases: training and classification. In the training phase, an eigenspace is established from the training samples using PCA and the training face images are mapped to the eigenspace for classification. In the classification phase, an input face is projected to the same eigenspace and classified by an appropriate classifier.

Contrasting the PCA that encodes information in an orthogonal linear space and it is determined that PCA keeps a set of linear features that best reflects the variance in a high-dimensional data. But, it gives no guarantee that this feature set is sufficient for better classification. Generally, original or transformed high-dimensional dataset projected onto the sub-space, which has higher variance but the classification capacity is very poor. On the other hand, by using generalization of LDA [7] in the form of canonical variate [15], the dataset is projected onto the sub-space, it shows lower variance, but classification rate is high. Canonical variates combine the class of each data item as well as the features that are estimating good set of features. Face images

Dakshina Ranjan Kisku is with the Department of Computer Science and Engineering, Dr. B. C. Roy Engineering College, Durgapur–713205, India (phone: +91-9732111234; e-mail: drkisku@gmail.com).

Hunny Mehrotra is with the Department of Computer Science and Engineering, National Institute of Technology Rourkela, Rourkela – 769008, India (e-mail: hunny@iitk.ac.in).

Jamuna Kanta Sing is with the Department of Computer Science and Engineering, Jadavpur University, Kolkata-700032, India (e-mail: jksing@ieee.org).

Phalguni Gupta is with the Department of Computer Science and Engineering, Indian Institute of Technology Kanpur, Kanpur-208016, India (e-mail: pg@cse.iitk.ac.in).

have many nonlinear characteristics that are not addressed by the only linear analysis methods discussed earlier, such as variations in illumination (outdoor lighting vs. indoor fluorescents), pose (standing straight vs. leaning over) and expression (smile vs. frown). For the well representation of feature space, a combine approach of representation of Gabor wavelets [12-13] and canonical covariate is proposed that construct strong Gabor canonical face responses that avoids the drawback of accurate localization of facial landmarks. At each feature point, feature vectors are extracted using the outputs of multi-scale and multi-orientation 2D Gabor wavelets.

In this paper, a face recognition based on Gabor wavelet transform is presented that convolves with face images of multiple views and facial expressions. Convolutions produce high dimensional feature vectors. In subsequent process, Gabor faces are encoded by canonical covariate to reduce the high dimensionality of the input feature spaces into the lower dimensional canonical covariates spaces. For recognition task, SVM classifier [11] is used. Our results show that Gabor canonical based representation of face images significantly improves the recognition performance over the methods presented in the literatures. It also observes that dimensionality reduction of feature vectors using canonical covariate can be use without loss of significant information in the feature spaces.

The rest of the paper is organized as follows. In the next section, Gabor wavelets representation for face images is introduced. In Section 3, canonical covariate for the sake of dimensionality reduction of high dimensional feature vectors is described. Section 4 describes Gabor-canonical covariate face representation of Gabor responses of multiview rotated faces. In Section 5, the overview of SVM for classification of face images is presented. Section 6 introduces the face database used in experimentation and describes the experiments carried out, and the results obtained. Conclusions are drawn in Section 7.

## II. OVERVIEW OF GABOR WAVELET

Gabor wavelet [12-13] has extensively been studied in biometrics, such as in face, fingerprint and palmprint biometrics. Due to its strong representation capability, Gabor wavelet transform is still be a feature extraction tool for some pattern recognition and biometric applications. Fundamentally, 2D Gabor filter [12-13] refers a linear filter whose impulse response function defines as the multiplication of harmonic function and Gaussian function. The Gaussian function is modulated by a sinusoid function. In this regard, the convolution theorem states that, the Fourier transform of a Gabor filter's impulse response is the convolution of the Fourier transform of the harmonic function and the Fourier transform of the Gaussian function. Gabor function is a non-orthogonal wavelet and it can be specified by the frequency of the sinusoid $\omega = 2\pi f$ and the standard deviations of $\sigma_x$ and $\sigma_y$. The 2D Gabor wavelet Filter can defines as follows

$$g(x, y : f, \theta) = \exp(-\frac{1}{2}(\frac{(x\sin\theta + y\cos\theta)^2}{\sigma_x^2} + .... \\ \frac{(x\cos\theta - y\sin\theta)^2}{\sigma_y^2})\cos(2\pi f(x\sin\theta + y\cos\theta))) \quad (1)$$

where f is the frequency of the sinusoidal plane wave along the direction $\theta$ from the x-axis, $\sigma_x$ and $\sigma_y$ specify the Gaussian envelop along x-axis and along y-axis, respectively. This can be used to determine the bandwidth of the Gabor filter. For the sake of experiment, 180 dpi gray scale face image with the size of 200 × 220 has been used. Along with this, 40 spatial frequencies are used for $f = \pi/2^i$, (i = 1,2,…,5) and $\theta = k\pi/8, (k = 1,2,...,8)$. For Gabor face representation, face image convolves with the Gabor wavelet transform for capturing substantial amount of variations among face images in the spatial locations. Gabor wavelet transforms with five frequencies and eight orientations use for generation of 40 spatial frequencies and for Gabor face extraction. In practice, Gabor face responses having very long representation vectors and the dimension of Gabor feature vector is prohibitively large.

The proposed work uses multiview face images for robust and invariant face recognition, in which any profile or frontal view query face can match to the database face image with a standard methodology of face verification strategy. First, the face images convolve with Gabor wavelet transforms and the convolution generates 40 spatial frequencies in the neighborhood regions of the current spatial pixel point. For the face image of size 200 × 220, 1760000 spatial frequencies generate. Infact, the huge dimension of Gabor responses could cause the performnce to be degrates and matching would be slow. Due to huge dimensionality of Gabor responses, dimensionality reduction operation is performed using canonical covariate.

In the next section, dimensionality reduction of gabor faces using canonical covariate is discussed.

## III. DIMENSIONALITY REDUCTION BY CANONICAL COVARIATE

The aim of dimensionality reduction of high dimensional features is to obtain a reduced set of features that best represents and reflects the relevance of the original feature set. The Gabor wavelet feature representation originated with very high dimensional space. It is necessary to reduce the high dimensional feature space to a low dimensional representation by selecting relevant and important features from the feature space. In the proposed work, canonical covariate [15] based holistic appearance technique is used to select the significant features from the Gabor face responses, and hence to reduce the high dimensional data.

In practice, Linear Discriminant Analysis (LDA) [7] can be used to find a single projection that can optimally separate the two classes in the distribution feature space, so that the within class distance would be maximized and between-cluster distance would be minimized.

The generalization of LDA [7], i.e., the canonical variate [15] used to project a dataset onto the sub-space and it shows lower variance, but classification probability is very high. Canonical variates combine the class of each data item as well as the features that are estimating good set of features.

To construct canonical variate representations for gabor face responses, let assume that there has a set of gabor face responses of C classes. Each class contains $n_k$ gabor responses and a gabor response from the $k^{th}$ class is $g_{k,i}$ for $i \in \varepsilon\{1,2,...,n_k\}$. Also assume that, the $C_j$ class has mean $\mu_j$ and there are d features (each $g_i$ is d-dimensional vectors). It can written $\overline{\mu}$ for the mean of the class means, that is

$$\overline{\mu} = \frac{1}{C}\sum_{j=1}^{C}\mu_j \quad (2)$$

and

$$\beta = \frac{1}{C-1}\sum_{j=1}^{C}(\mu_j - \overline{\mu})(\mu_j - \overline{\mu})^T. \quad (3)$$

For illustration, ß denote the variance of the class means. In the generalized case, each class has the identical covariance $\sum$, and that has the full rank. In order to obtain a set of axes where the feature points have grouped into some clusters belonging to a particular class group together tightly and the distinct classes have widely separated. This involves finding a set of features that maximizes the ratio of the separation (i.e., variance) between the class means to the variance within each class. The separation between the class means is typically referred to as the between-class variance, and the variance within a class is typically referred to as the within class variance.

Let us consider, each class has the identical covariance $\sum$, which is either known or estimated as

$$\sum = \frac{1}{N-1}\sum_{s=1}^{C}\left\{\sum_{i=1}^{n_c}(g_{s,i} - \mu_s)(g_{s,i} - \mu_s)^T\right\}. \quad (4)$$

From the Equations (3) and (4), the unit eigenvectors of UV can be defined as

$$UV = \sum\nolimits^{-1}\beta = [ev_1, ev_2,...,ev_m] \quad (5)$$

where each $ev_i(ev_i | i = 1,2,...,m)$ denote the eigenvalues and the dimension m denotes the size of the eigenvalues and ev1 has the largest eigen-value, gives a set of linear gabor features that best separates the class means. Projection onto the basis $\{ev_1, ev_2,...,ev_k\}$ provides the k-dimensional set of linear features that best separates the class means.

IV. GABOR CANONICAL FACE REPRESENTATION

When a high dimensional Gabor face image is projected onto a sub-space using canonical covariates, it shows lower variance in high dimensional data, but the classification rate of faces is high. Canonical covariate is used to estimate a significant set of features from the high dimensional Gabor filter representation. Then combine this estimated feature set with the class of each data item. Finally, reduced set of Gabor-canonical faces are produced. Due to useful characteristics of Gabor responses, its representation is more robust and effective against varying brightness and contrast in the face image.

With this representation, the Gabor canonical face responses remove some shortcomings. Due to the pose, illumination, expressions, etc, some major changes and shortcomings often occurs in the face image. For robust and efficient classification of face images, the Gabor filter responses project onto another sub-space using canonical covariate based on the principal axis in terms of linear features. Then by using Equation (4) and Equation (5), a reduced set of canonical covariates of Gabor responses is generated.

V. SUPPORT VECTOR MACHINES FOR FACE CLASSIFICATION

The proposed work use support vector machines [11] to solve the problem of classifying faces. The training problem can be formulated as separating hyper-planes that maximizes the distance between closest points of the two classes. In practice, this is determined through solving quadratic problem. The SVM has a general form of the decision function for N data points $\{x_i, y_i\}_{i=1}^{N}$, where $x_i \in \mathbb{R}^n$ the i-th input data is, and $y_i \in \{-1,+1\}$ is the label of the data. The SVM approach aims at finding a classifier of form:

$$y(x) = sign\left[\sum_{i=1}^{N}\alpha_i y_i K(x_i, x) + b\right] \quad (6)$$

Where $\alpha_i$ are positive real constants and $b$ is a real constant, in general, $K(x_i, x) = \langle\phi(x_i), \phi(x)\rangle$ is known as kernel function where $\langle\cdot,\cdot\rangle$ inner product is, and $\phi(x)$ is the nonlinear map from original space to the high dimensional space. The kernel function can be various types. The linear function is defined as $K(x, y) = x \cdot y$; the radial basis function (RBF) kernel function is $K(x,y) = \exp(-\frac{1}{2\sigma^2} \|x-y\|^2)$; and the polynomial kernel function is $K(x,y) = (x \cdot y + 1)^n$. SVM can be designed for either binary classification or multi-classification. For the sake of experiment, binary classification approach is used. In "one-vs-one" binary classification, the decision function (6) can be written as

$$f(x) = sign(\omega \cdot x - b) \qquad (7)$$

where ω (inner product of weight vector) is obtained from the following equation

$$\omega = \sum_i y_i \alpha_i x_i \qquad (8)$$

The input feature vector x and weight vector ω determines the value of f(x). During classification, the input features with the largest weights correspond to the most discriminative and informative features. Therefore, the weights of the linear function can be used as final classification criterion for binary decision of multiview faces.

## VI. EXPERIMENTAL RESULTS

In this section, the experimental results of the proposed method on UMIST face database [14], [20] describe. UMIST face database consists of 564 face images of 20 distinct persons or subjects. Faces in the database cover range of poses from profile to frontal view. Each subject covers a mixed range of race, sex and appearance. Such as different expressions, illuminations, glasses/no glasses, beard/no beard, different hair style etc. Some images are shown in Figure 2 for a single subject.

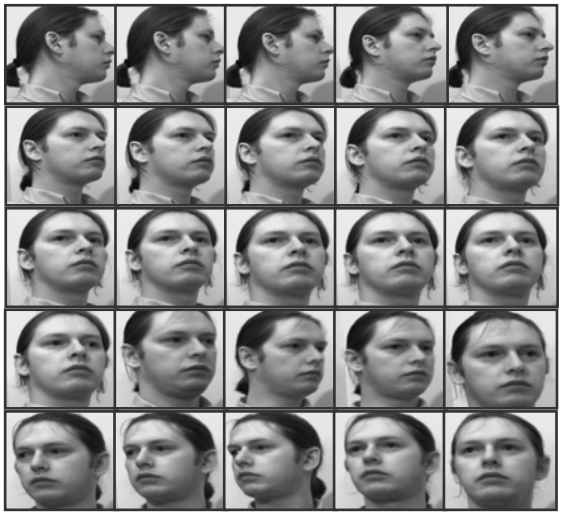

Fig. 1 Sample face images from UMIST database for a single subject are shown.

The experiment is accomplished in three steps. In the first step feature representation is done by Gabor wavelet feature. Canonical Covariate performs dimensionality reduction of high dimensional Gabor wavelet features by finding significant linear features that best separates multiple classes efficiently. After feature reduction process, Support vector machine (SVM) classifier with two kernel functions does classification work of faces, namely, radial basis function and linear function kernels.

Apart from SVM classifier, that shows the superiority of the proposed method, classification task using K-nearest neighbor approaches with the Euclidean distance measure and cosine metric is illustrated.

Before feature representation of faces using Gabor wavelet, pre-processing operations perform in order to obtain the recognized face with most similar probability.

In order to get the normalized cropped faces; the face image registration done by semi-automatic approaches by removing unwanted facial features that affect the overall performance of the present system including hair, both the ears, etc. To detect the face portion that contains fore head, both eyes, nose, moving salient feature mouth, eye positions are localized manually. The essential parameters are computed from the eye coordinates and eye canthus positions (rotation, scaling and translation in the horizontal and vertical directions) are used to crop the facial part from the original image, which contains the important salient facial features.

In next step, face images photometrically normalize by histogram equalization to make uniform illumination effect over the cropped face image.

Face images are now ready to identity verification. The experiments are conducted on subset of face images taken from UMIST [14], [20] face database. For the sake of experiment, subset of image database is divided into two different sets of images. First set contains face images of frontal views. Test to be performed on first set, 6 frontal view faces are taken per subject. In the next phase of experiment, training of face images is accomplished with multiple profiles including frontal profile, left profile and right profile face images and match is performed with varying ranges of faces in terms of rotated face images. For the second set, 6 frontal views, 6 left rotated views and 6 right rotated views are taken for evaluation. In this paper, the problem of how to cope up with multiple frontal and profile rotated face images within a single experimental environment is focused. Relevant to these constraints, there are two approaches of differentiating several classes (users) efficiently by SVM. The approaches are such as binary classification between two classes and the 'one vs. all' multi-classification between several classes. In the proposed implementation, the first one is adopted. Using SVM the binary classification is accomplished between two classes. By using a kernel function the non-linear separable vector is mapped into a high-dimensional 'feature' space and thereby makes it separable. On both the data sets the experiments are performed by binary classification methodology. Experimental results computed from the first set shows that SVM outperforms other methods with radial basis function used as a kernel. The performance of SVM with linear function kernel also investigated and results are quite impressive. Other two nearest neighbor methods namely, Euclidean distance metric and cosine are also compared with SVM with two kernel functions. For the first set of face images, that contains frontal profiles, the DET (detection error trade-off) curve and the ROC (receiver operating characteristics) curves are shown in Figure 2 and Figure 3, respectively. FAR (False accept rate) and FRR (False reject rate) of all the methods which are proposed in this paper are shown in Table 1.

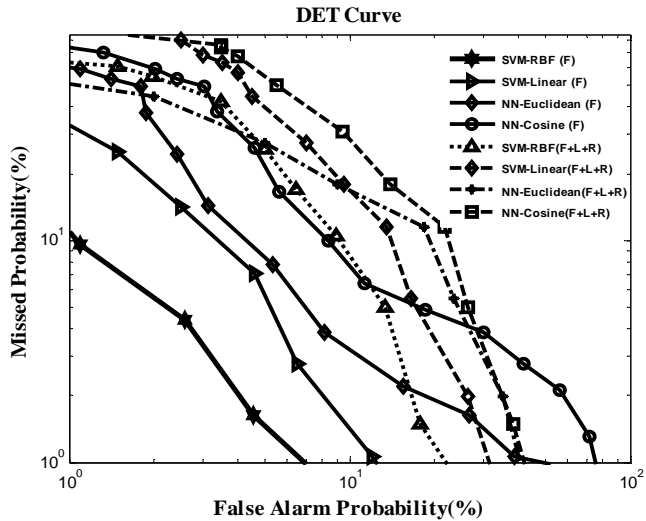

Fig. 2 DET curves for different proposed methods are shown.

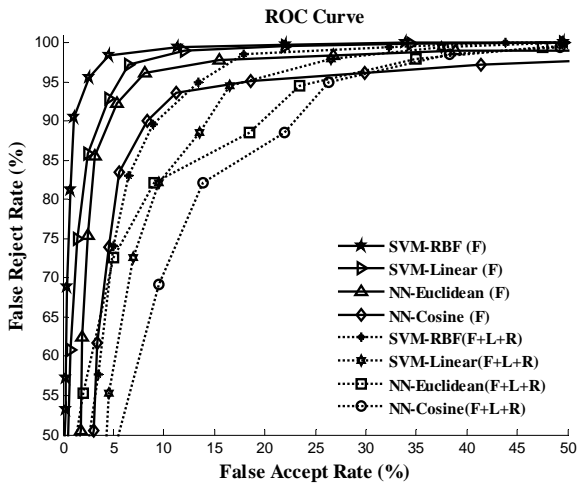

Fig. 3 Receiver operating characteristics curves for different proposed methods are shown.

In order to perform the evaluation on second set of face images SVM is also being trained with multiple profile face images viz. frontal view, rotated profiles views, rather than frontal view faces. In this experiment, the SVM is trained with multiple views of reduced Gabor representation faces. On the second set of face images the SVM is again outperform other methods with radial basis function kernel. The DET curves and ROC curves are shown in Figure 2 and Figure 3, respectively, in consideration with multiple views of faces. FAR and FRR are shown in Table 1.

TABLE I
THE TABLE SHOWS FRR, FAR, AND EER COMPUTED FROM DIFFERENT METHODS AND EVALUATED ON TWO SETS OF FACE DATASETS, IN WHICH FIRST SET CONTAINS FRONTAL VIEW FACES AND THE SECOND SET CONTAINS OF MULTIPLE VIEW FACES. "F" STANDS FOR FRONTAL VIEW FACE, "L" STANDS FOR LEFT VIEW FACE AND "R" STANDS FOR RIGHT VIEW FACE.

| Technique and methods | FRR (%) | FAR (%) | Recognition Rate (%) | EER (%) |
|---|---|---|---|---|
| SVM-RBF (F) | 2.0192 | 4.5168 | 96.732 | 3.268 |
| SVM-Linear (F) | 3.2303 | 6.9674 | 94.9211 | 5.099 |
| NN-Euclidean (F) | 3.4012 | 8.4332 | 94.082 | 5.9172 |
| NN-Cosine (F) | 7.1335 | 11.0465 | 90.91 | 9.09 |
| SVM-RBF (F+L+R) | 7.3453 | 13.9802 | 89.3372 | 10.6627 |
| SVM-Linear (F+L+R) | 11.00 | 14.4328 | 87.2876 | 12.7164 |
| NN-Euclidean (F+L+R) | 13.9 | 17.0204 | 84.5398 | 15.4602 |
| NN-Cosine (F+L+R) | 15.0204 | 18.8104 | 83.0843 | 16.9154 |

VII. CONCLUSION

The aim of the paper is to reports a robust face recognition method that copes with multiview faces with facial expressions efficiently. Due to high dimensionality of Gabor face responses, the reduced set of features obtained by applying generalization of LDA in form of canonical covariate. The reduced set of Gabor features identifies themselves as significant set of features. For classification of faces support vector machine is used to classify the faces in binary classification pattern. Performance of the proposed system estimated as a robust face recognition system when it compared with the other methods presented in this paper and with methods presented in the literatures. Reported results confirm the efficacy of the proposed scheme while the method tested on the UMIST dataset with quite complex multiview faces. The performance of RBF kernel based SVM is much better than that of the linear kernel based SVM classifier and the K-nearest neighbor classifiers. The ROC-curves show that the robustness and reliability of the recognition performance rectified and increases with proper selection feature representation scheme for multiview faces. The performance largely dependent on three factors: the dataset to be uses, combined feature representation scheme and the right classifier. It is worthless to use poor representation model and weak classifier that resulted to be decrease in overall accuracy and recognition performance. Because of the reported results, it is worth devoting further experimental investigations to understand the behavior of canonical covariate in order to integrate with appearance model. Irrespective of certain difficulties present in captured faces such as illumination changes, pose changes, presence of occlusion, expression changes, etc; the present system can combat with these problem and classify faces efficiently.

REFERENCES

[1] J. Y. Gan, and Y. W. Zhang, "A new approach for face recognition based on singular value features and neural networks," *Acta Electronica Sinica*, vol. 32, no.1, pp. 56 – 58, 2004.
[2] J. Y. Gan, Y. W. Zhang, and S. Y. Mao, "Adaptive principal components extraction algorithm and its applications in the feature extraction of human face," *Acta Electronica Silica*, vol. 30, no. 7, pp. 1013 - 1016, 2002.
[3] M. Dai, and M. Q. Zhou, "On automatic human face recognition," *Advances Biometrics*, vol. 1, pp. 41 - 48, 2003.
[4] M. Turk, and A. Pentland, "Eigenfaces for recognition", *Journal of Cognitive Neuroscience*, vol. 3, no. 1, pp. 71 – 86, 1991.
[5] M. Turk, and A. Pentland, "Face recognition using eigenfaces", *Proceeding of the IEEE Conference on Computer Vision and Pattern Recognition*, 1991, pp. 586 – 591.
[6] P. Belhumeur, J. Hespanha, and D. Kriegman, "Eigenfaces vs. fisherfaces: Recognition using class specific linear projection", *Proceeding of the Fourth European Conference on Computer Vision*, vol. 1, 1996, pp. 45 – 58.
[7] A. Martinez, and A. Kak, "PCA versus LDA", *IEEE Transaction on Pattern Analysis and Machine Intelligence*, vol. 23, no. 2, pp. 228 – 233, 2001.
[8] G. W. Cottrell, and M. K. Fleming, "Face recognition using unsupervised feature extraction," *Proceedings of the International Conference on Neural Network*, 1990, pp. 322 – 325.
[9] M. S. Bartlett, J. R. Movellan, and T. J. Sejnowski, "Face recognition by independent component analysis," *IEEE Transaction on Neural Networks*, vol. 13, no. 6, pp. 1450 – 1464, 2002.
[10] M. H. Yang, "Kernel eigenfaces vs. kernel fisherfaces: Face recognition using kernel methods," Proceedings of the Fifth IEEE International Conference on Automatic Face and Gesture Recognition, 2002, pp. 215 – 220.
[11] C. J. C. Burges, "A tutorial on support vector machines for pattern recognition," *Data Mining and Knowledge Discovery*, vol. 2, no. 2, pp. 121–167, 1998.
[12] J. G. Daugman, "Complete discrete 2-D gabor transforms by neural networks for image analysis and compression", *IEEE Transaction on Acoustic, speech and signal processing*, vol. 36, pp.1169 – 1179, 1998.
[13] T. S. Lee, "Image representation using 2D gabor wavelets", *IEEE Transaction on Pattern Analysis and Machine Intelligence*, vol. 18, pp.959 – 971, 1996.
[14] http://images.ee.umist.ac.uk/danny/database.html.
[15] H. Hotelling, "Relations between two sets of variates", *Biometrika*, vol. 28, pp. 321– 377, 1936.
[16] D. J. Beymer. "Face recognition under varying pose," MIT AI Lab, Technical Report, 1993.
[17] A. Pentland, B. Moghaddam, and T. Starner, "View-based and modular eigenspaces for face recognition", Proceedings of the International Conference on Computer Vision and Pattern Recognition. 1994.
[18] V. Blanz, and T. Vetter, "A morphable model for the synthesis of 3D faces," Proceedings of the International Conference SIGGRAPH, 1999, pp. 187 – 194.
[19] J. Lu, K. N. Plataniotis, and A. N. Venetsanopoulos, "A kernel machine based approach for multi-view face recognition," IEEE International Conference on Image Processing, 2002, pp. 265 – 268.
[20] A. Rattani, D. R. Kisku, A. Logario, and M. Tistarelli, "Facial template synthesis using SIFT features," IEEE Workshop on Automatic Identification Advanced Technologies, 2007, pp. 69 – 73.

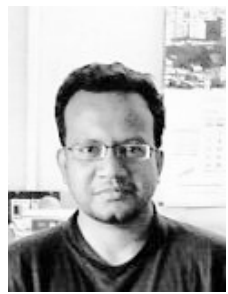

**Dakshina Ranjan Kisku** was born on July 1, 1977 in Durgapur, India. He received the B.Engg. and M.Engg. degrees in Computer Science and Engineering from Jadavpur University, Kolkata, India, in 2001 and 2003, respectively. Currently, he is pursuing Ph.D. in Computer Science and Engineering at the Jadavpur University. His research interests include computer vision, pattern recognition and biometrics.

From March 2006 to March 2007, he was a Researcher at the Computer Vision Laboratory, University of Sassari, Italy. He also worked as a Research Associate at Indian Institute of Technology Kanpur, India in 2005. From August 2003 to August 2005, he was a Lecturer in Computer Science and Engineering Department at the Asansol Engineering College, India.

Mr. Kisku is a member of IEEE and IET. He has published several research papers on biometrics in leading peer-reviewed conferences. Mr. Kisku also worked as a reviewer for several conferences.

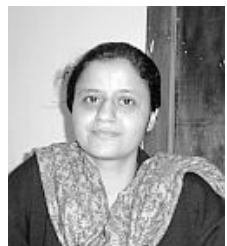

**Hunny Mehrotra** was born on November 20, 1983 in Kanpur, India. She received the MCA degree in Computer Application in 2004. Currently, she is pursuing M.Tech (Research) in Computer Science and Engineering at National Institute of Technology, Rourkela, India. Her research interests include biometrics, image processing, and pattern recognition.

From 2004 to 2007, she was a Research Associate at Indian Institute of Technology Kanpur, India.

Ms. Mehrotra has published many research papers in peer-reviewed conferences and journals. Ms. Mehrotra also worked as a reviewer for several conferences.

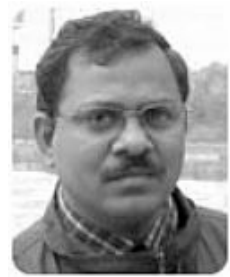

**Dr. Jamuna Kanta Sing** was born in 1974 in West Bengal, India. He received the B.Engg degree in Computer Science and Engineering from Jadavpur University, Kolkata, India, in 1992 and M.Tech degree in Computer and Information Technology from Indian Institute of Technology Kharagpur, India, in 1994. He earned the Ph.D. degree from Jadavpur University in 2006. His research interests include image processing, medical imaging, pattern recognition, artificial neural networks, computer vision, etc.

Currently Dr. Sing working as a Associate Professor in Computer Science and Engineering Department at Jadavpur University, Kolkata, India. He was a Scientist at National Physical Laboratory in 1995. He then joined as a lecturer at Jadavpur University in 1996 and was at the same position till 2002. From 2003 to 2006, he worked as a Senior Lecturer. In 2005, he was awarded BOYSCAST fellowship for pursuing postdoctoral research at the University of IOWA and he continued his research study until 2007. He is also serving as a Chairman of the GOLD affinity group of IEEE Kolkata section since last three years. He was a Principal Investigator for many DST, AICTE sponsored research projects on diverse fields of image processing and medical imaging.

He is member of IEEE. His contribution includes many significant works, which have published in many peer-reviewed conferences and journals. He has been working as a reviewer for several conferences and journals.

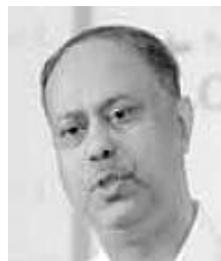

**Prof. Phalguni Gupta** received his Ph.D. degree from Indian Institute of Technology, Kharagpur, India in 1986. He works in the field of data structures, sequential algorithms, parallel algorithms, on-line algorithms and image analysis.

From 1983 to 1987, Prof. Gupta was in the Image Processing and Data Product Group of the Space Applications Centre (ISRO), Ahmedabad, India and was responsible for software for correcting image data received from Indian Remote Sensing Satellite. In 1987, he joined the Department of Computer Science and Engineering, Indian Institute of Technology Kanpur, India. Currently he is a Professor in the department. He is involved in several research projects in the area of Biometric Systems, Mobile Computing, Image Processing, Graph Theory and Network Flow.

He is member of ACM and IEEE. He has made many significant contributions to parallel algorithms, on-line algorithms and image processing and published many research papers in conferences and journals.